\documentclass[11pt,a4paper]{article}
\usepackage[hyperref]{acl2021}
\usepackage{times}
\usepackage{latexsym}
\usepackage{amssymb,mathrsfs,amsmath}

\usepackage{microtype}

\usepackage{times}
\usepackage{url}
\usepackage{latexsym}
\usepackage{url}
\usepackage{xcolor}
\usepackage{booktabs}

\usepackage{algorithm}
\usepackage{algorithmicx}
\usepackage{algpseudocode}

\floatname{algorithm}{Algorithm}

\usepackage{graphicx}
\usepackage{subfigure}

\title{Why Machine Reading Comprehension Models Learn Shortcuts?}

\aclfinalcopy

\author{
    Yuxuan Lai, 
    Chen Zhang, 
    Yansong Feng\thanks{\;\;Corresponding author.}~~,  
    Quzhe Huang, \and
    Dongyan Zhao \\
    Wangxuan Institute of Computer Technology, Peking University, China\\
    The MOE Key Laboratory of Computational Linguistics, Peking University, China\\
    {\tt \{erutan, zhangch, fengyansong, huangquzhe, zhaody\}}\\
    {\tt @pku.edu.cn} \\
}

\date{}

\begin{document}
\maketitle
\begin{abstract}
Recent studies report that many machine reading comprehension (MRC) models can perform closely to or even better than humans on benchmark datasets.
However, existing works indicate that many MRC models may learn \textit{shortcuts} to outwit these benchmarks, but the performance is unsatisfactory in real-world applications.
In this work, we attempt to explore, instead of the expected comprehension skills, why these models learn the \textit{shortcuts}.
Based on the observation that a large portion of questions in current datasets have \textit{shortcut} solutions, we argue that larger proportion of \textit{shortcut} questions in training data make models rely on \textit{shortcut} tricks excessively. 
To investigate this hypothesis, we carefully design two synthetic datasets with annotations that indicate whether a question can be answered using \textit{shortcut} solutions.
We further propose two new methods to quantitatively analyze the learning difficulty regarding \textit{shortcut} and challenging questions, and revealing the inherent learning mechanism behind the different performance between the two kinds of questions. 
A thorough empirical analysis shows that MRC models tend to learn \textit{shortcut} questions earlier than challenging questions, and the high proportions of \textit{shortcut} questions in training sets hinder models from exploring the sophisticated reasoning skills in the later stage of training.
\end{abstract}

\section{Introduction}
\label{intro}

The task of machine reading comprehension (MRC) aims at evaluating whether a model can understand natural language texts by answering a series of questions.
Recently, MRC research has seen considerable progress in terms of model performance, and many models are reported to approach or even outperform human-level performance on different benchmarks. These benchmarks are designed to address challenging features, such as evidence checking in multi-document inference \cite{hotpot}, co-reference resolution \cite{dasigi-etal-2019-quoref}, dialog understanding \cite{coqa}, symbolic reasoning \cite{drop}, and so on. 

\begin{figure}[t]
    \centering
    \includegraphics[width=0.5\textwidth]{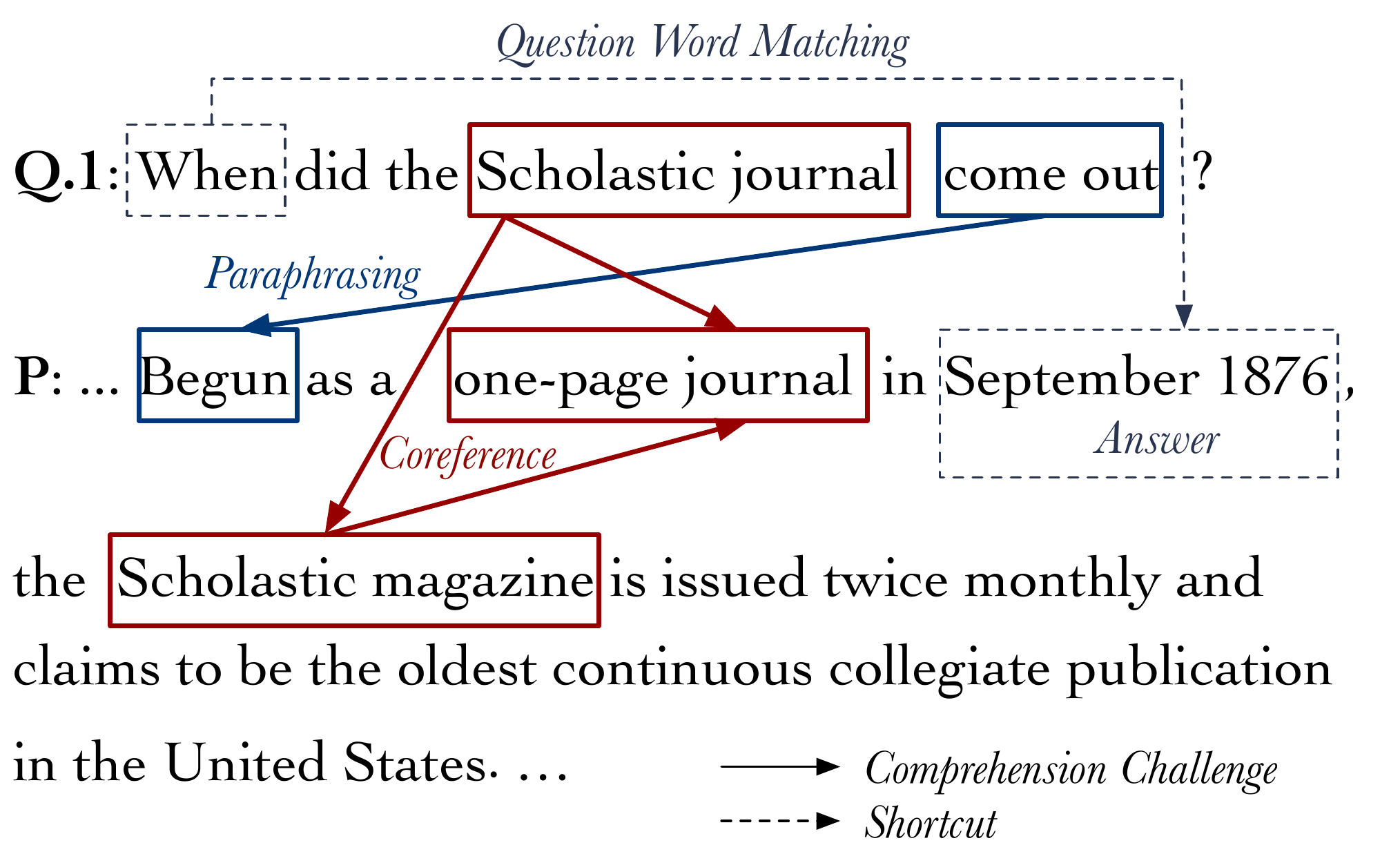}
    \caption{\label{ex1} An illustration of \textit{shortcuts} in Machine Reading Comprehension. 
    \textbf{P} is an excerpt of the original passage.
    }
\end{figure}

However, recent analysis indicates that many MRC models unintentionally learn \textit{shortcuts} to trick on specific benchmarks, while having inferior performance in real comprehension challenges \cite{sugawara2018makes}.
For example, when answering \textbf{Q.1} in Figure~\ref{ex1}, we expect an MRC model to understand the semantic relation between \textit{come out} and \textit{begun}, and output the answer, \textit{September 1876}, by bridging the co-reference among \textit{Scholastic journal}, \textit{Scholastic magazine} and \textit{one-page journal}.
In fact, a model can easily find the answer without following the mentioned reasoning process, since it can just recognize \textit{September 1876} as the only time expression in the passage to answer a \textit{when} question. We consider such kind of tricks that use \textit{partial} evidence to produce, perhaps unreliable, answers as \textbf{\textit{shortcuts}} to the expected comprehension challenges, e.g., \textit{co-reference resolution} in this example.
The questions with \textit{shortcut} solutions are referred to as \textit{\textbf{shortcut questions}}.
For clarity, a model is considered to have \textit{\textbf{learned shortcuts}} when it relies on those tricks to obtain correct answers for most \textit{shortcut} questions while performing worse on questions where challenging skills are necessary. 

Previous works have found that, relying on \textit{shortcut} tricks, models may not need to pay attention to the critical components of questions and documents \cite{mudrakarta2018did} in order to get the correct answers. 
Thus, many current MRC models can be either vulnerable to disturbance \cite{jia2017adversarial}, or lack of flexibility to question/passage changes \cite{sugawara2020}.
These efforts disclose the impact of \textit{shortcut} phenomenon on MRC studies.
However, concerns have been raised on \textbf{why MRC models learn these \textit{shortcuts} while ignoring the designed comprehension challenges.}

To properly investigate this problem, our first obstacle is that there are no existing MRC datasets that are labeled whether a question has \textit{shortcut} solutions.
This deficiency makes it hard to formally analysis how the performance of a model is affected by the \textit{shortcuts questions}, and almost impossible to examine whether the model correctly answers a question via \textit{shortcuts}. 
Secondly, previous methods disclose the \textit{shortcut} phenomenon by analyzing the model outputs through a series of carefully designed experiments, but fail to explain how the MRC models learn the \textit{shortcuts} tricks.
We need new methods to help us quantitatively investigate the learning  mechanisms that make the difference when MRC models learn to answer the \textit{shortcuts} questions and questions that require challenging reasoning skills.

In this work, we carefully design two synthetic MRC datasets to support our controlled experimental analysis.
Specifically, in these datasets, each (passage, question) instance has a \textit{shortcut} version paired with a challenging one where complex comprehension skills are required to answer the question. 
Our construction method ensures that the two versions of questions are as close as possible, in terms of style, size, and topics, which enable us to conduct controlled experiments regarding the necessary skills to obtain answers.
We design a series of experiments to quantitatively explain how \textit{shortcut} questions affect MRC model performance and how the models learn these tricks and challenging skills during the training process.
We also propose two evaluation methods to quantify the learning difficulty of specific question sets.
We find that \textit{shortcut} questions are usually easier to learn, and the dominant gradient-based optimizers drive MRC models to learn the \textit{shortcut} questions earlier in the learning process.
The priority of fitting \textit{shortcut} questions hinders models from exploring sophisticated reasoning skills in later stage of training.
Our code and datasets can be found in \url{https://github.com/luciusssss/why-learn-shortcut}

To summarize, our main contributions are the following: 
1) We design two synthetic datasets to study two commonly seen \textit{shortcuts} in MRC benchmarks, \textit{question word matching} and \textit{simple matching}, against a challenging reasoning pattern \textit{paraphrasing}.
2) We propose two simple methods as a probe to help investigate the inherent learning mechanism behind the different performance on \textit{shortcut} questions and challenging ones. 
3) We conduct thorough experiments to quantitatively explain the behaviors of MRC models under different settings, and show that the proportions of \textit{shortcut} questions greatly affect model performance, which may hinder MRC models from learning sophisticated reasoning skills.

\begin{figure*}[ht]
    \centering
    \includegraphics[width=1\textwidth]{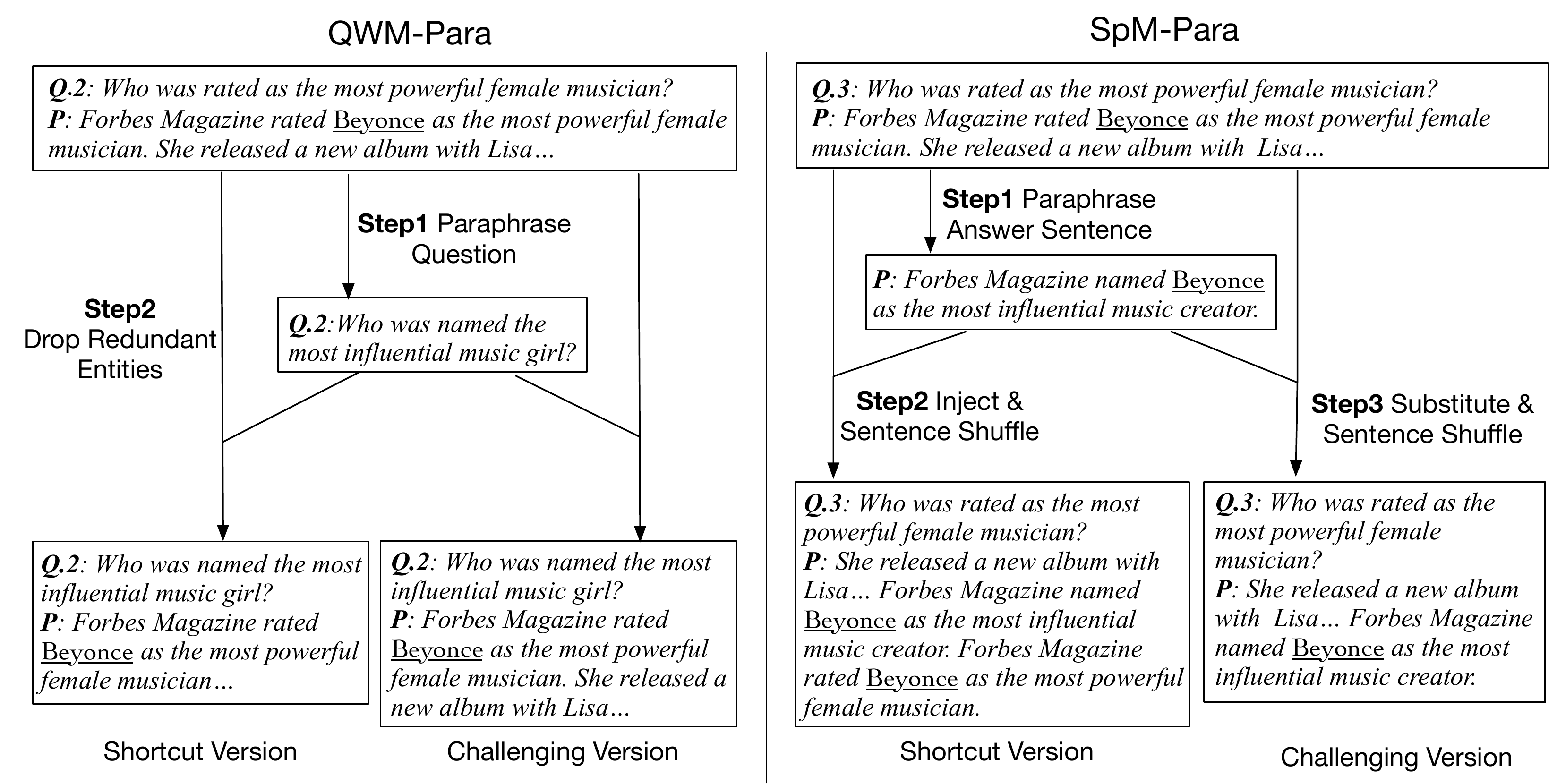}
    \caption{An illustration of how the instances in the synthetic datasets are constructed from original SQuAD data. Each instance has a \textit{shortcut} version paired with a challenging version where comprehension skills are necessary. 
    \label{fig:lct}
    }
\end{figure*}

\section{Synthetic Dataset Construction}
\label{syntheticdataset}

To study the impact of \textit{shortcut} questions in model training, we require the datasets to be annotated with whether each question has \textit{shortcut} solutions, or can only be answered via complex reasoning. 
However, none of existing MRC datasets have such annotations.  
We thus design two synthetic datasets where it is known whether \textit{shortcut} solutions exist for a question.

More importantly, we need to conduct controlled experiments and ensure, for each question, the existence of \textit{shortcuts} solutions is the only independent variable.
The extraneous variables, such as sizes of datasets, topics, answer types, and even the vocabulary, should be controlled relatively steady.
Thus, in our designed datasets, each entry has a \textit{shortcut} version instance and a challenging version.
The question of the \textit{shortcut} version can be correctly solved by a certain \textit{shortcut} trick, while an expected comprehension skill is required to deal with the challenging version.
Note that we expect the two versions of questions are as close as possible so that we can switch between the two versions freely  while maintaining other factors relatively steady.

In this work, we focus on \textit{paraphrasing} (\textit{\textbf{Para}}) as the complex reasoning challenge, since it widely exists in many recent MRC datasets \cite{newsqa,coqa,boolq}.
The paraphrasing challenge requires MRC models to identify the same meaning represented in different words.
Regarding the \textit{shortcut} tricks,  we study two typical kinds: \textit{question word matching} (\textbf{\textit{QWM}}) and \textit{simple matching} (\textbf{\textit{SpM}}) \cite{sugawara2018makes}.
For \textit{QWM}, MRC models can simply obtain an answer phrase by recognizing the expected entity type confined by the \textit{wh-}question words of question $Q$.
For \textit{SpM}, a model can find the answers by identifying the word overlap between answer sentences and the questions.

\begin{algorithm}[t]
\centering
\small
  \caption{\label{algo_qwm}Construction of \textit{QWM-Para}}
  \begin{algorithmic}[1] 
      \Require SQuAD 
      \Ensure \textit{QWM-Para}
      \State \textit{QWM-Para} $\leftarrow \emptyset$
      \For{each instance $(Q, P)$ in SQuAD}
        \If{$Q$ does not start with \textit{who}, \textit{when}, \textit{where}}
          \State Discard this instance.
        \EndIf
        \If{the answer sentence contains other entities matching the question word}
          \State Discard this instance.
        \EndIf
        \State Use back translation to paraphrase $Q$, obtain $Q_p$
        \If{the non-stop-word overlap rate between $Q_p$ and the answer sentence \textgreater 25\%}
            \State Discard the instance.
        \EndIf
      \State  Delete sentences in passage $P$ that does not contain the golden answer but containing other entities matching the question word, note the modified passage as $P_s$.
    \State $I_s \leftarrow$ the \textit{shortcut} instance version $\left( Q_p, P_s \right)$
    \State $I_c \leftarrow$ the \textit{challenging} instance version $\left( Q_p, P \right)$
    \State Append the pair of questions, $\left( I_s, I_c \right)$, to \textit{QWM-Para}.
      \EndFor
  \end{algorithmic}
\end{algorithm}

{\textbf{\textit{QWM-Para} Dataset:~}} 
As elaborated in Algorithm~\ref{algo_qwm}, given an original instance  $(Q, P)$ from SQuAD \cite{squad}, we paraphrase the question $Q$ in $Q_p$ to embed the \textit{paraphrasing} challenge, and derive the corresponding \textit{shortcut} version by dropping the sentences  containing other entities with the matched type according to the question words from the given passage. 

An example is shown in the left of Figure~\ref{fig:lct}.
In the challenging version of \textbf{Q.2}, both \textit{Beyonce} and \textit{Lisa} are person names which match the question word \textit{who}.
Thus, one should at least recognize the paraphrasing relationship between \textit{named the most influential music girl} and \textit{rated as the most powerful female musician} to distinguish between the two names to infer the correct answer.
For the \textit{shortcut} version,  removing the sentence containing \textit{Lisa} from the passage,
which is also of the expected answer type \textit{person} indicated by the question word \textit{who}, would help a model easily get the correct answer, \textit{Beyonce}.

{\textbf{\textit{SpM-Para} Dataset:~}} 
As shown in Algorithm~\ref{algo_spm}, for instances from SQuAD, we paraphrase the answer sentences in the passage to embed the \textit{paraphrasing} challenge and obtain its challenging version.
We insert the paraphrased answer sentence in front of the original one in the passage to construct the corresponding \textit{shortcut} version, where a model can obtain the answers by either identifying the paraphrase in the passage or using the \textit{simple matching} trick via the original answer sentences.
We randomly shuffle all sentences in the passage to prevent models from learning the pattern of sentence orders in the
\textit{shortcut} version, i.e., there are two adjacent answer sentences in the passage.
Here, we assume the sentence-level shuffling operation will not affect the answers and solutions for most questions, since the supporting evidence is often concentrated in a single sentence. This can also be supported by \citet{sugawara2020}'s findings that  the performance of BERT-large~\cite{bert} on SQuAD only drops by around 1.2\% after sentence order shuffling. 

For example, in the \textit{shortcut} version of \textbf{Q.3} shown in the right of Figure~\ref{fig:lct}, MRC models can find the answer, \textit{Beyonce}, either from the matching context, \textit{rated as the most powerful female musician}, or via the paraphrased one, \textit{named as the most influential music girl}.
For the challenging version, only the paraphrased answer sentence is provided, thus, the paraphrasing skill is necessary.

\begin{algorithm}[t]
\centering
\small
  \caption{\label{algo_spm}Construction of \textit{SpM-Para}}
  \begin{algorithmic}[1] 
      \Require SQuAD
      \Ensure \textit{SpM-Para}
      \State \textit{SpM-Para} $\leftarrow \emptyset$
      \For{each instance $(Q, P)$ in SQuAD}
        \If{the non-stop-word overlap rate between $Q$ and the answer sentence $S$ \textless 75\%}
            \State Discard the instance.
        \EndIf
        \State Use back translation to paraphrase the answer sentence $S$ in $P$, obtain $S_p$.
        \If{the answer span no longer exists in $S_p$}
          \State Discard this instance.
        \EndIf
        \If{the non-stop-word overlap rate between $Q$ and  $S_p$ \textgreater 25\%}
            \State Discard the instance.
        \EndIf
      \State Replace $S$ in $P$ with $S_p$ and shuffle sentences, noted the modified passage as $P_c$.
      \State Append $S_p$ to $P$ and shuffle sentences, noted the modified passage as $P_s$.
      \State $I_s \leftarrow$ the \textit{shortcut} instance version $\left( Q, P_s \right)$
      \State $I_c \leftarrow$ the \textit{challenging} instance version $\left( Q, P_c \right)$
      \State Appen d the pair of questions, $\left( I_s, I_c \right)$, to \textit{SpM-Para}.
      \EndFor
  \end{algorithmic}
\end{algorithm}

\paragraph{Dataset Details}~
Our synthetic training and test sets are derived from the accessible training and development sets of SQuAD, respectively.
We adopt back translation to obtain paraphrases of texts \cite{dong2017learning}.  
A sentence is translated from English to German, then to Chinese, and finally back to English to obtain its paraphrased version.\footnote{We use Baidu Translate API (\url{http://api.fanyi.baidu.com}).}

The \textit{QWM-Para} dataset contains 7072 entries, each containing two versions of (question, passage) tuples, 6306/766 for training and testing, respectively. 
And for \textit{SpM-Para}, there are 8514 entries, 7562/952 for training and testing, respectively.

\paragraph{Quality Analysis}~ We randomly sample 20 entries from each training set of the synthetic datasets, manually analyzing their answerability. 
We find that 76/80 questions could be correctly answered. The unanswerable questions result from the wrong paraphrasing. 
Furthermore, among the answerable questions, the paraphrasing skill is necessary in 30 out of 36 questions in the challenging version, and 36 out of 40 questions of the \textit{shortcut} version  can be correctly answered via the corresponding \textit{shortcut} trick.

\begin{figure*}[t]
\centering
\includegraphics[width=1.\textwidth]{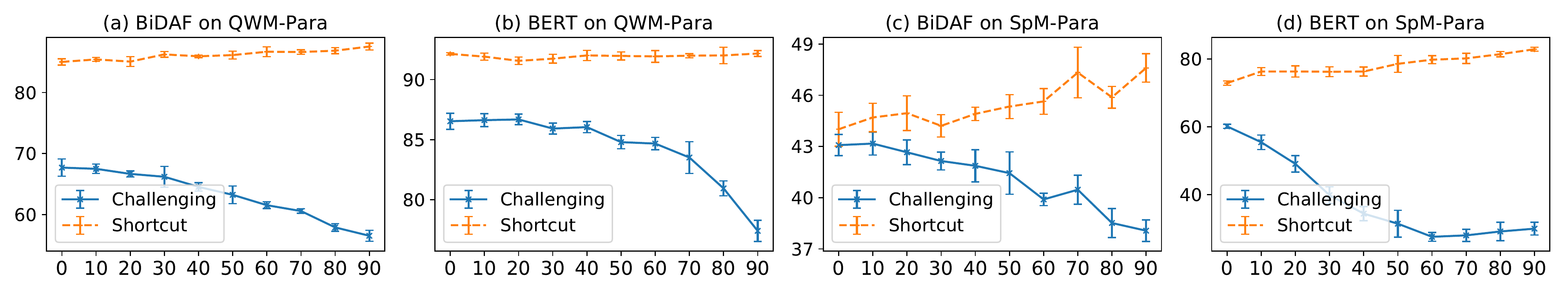}
\caption{\label{bidafbert_anstype}
F1 scores on challenging and \textit{shortcut} questions with different proportions of \textit{shortcut} questions in training.
The error bars represent the standard deviations of five runs.
}
\end{figure*}

\section{How the \textit{Shortcut} Questions Affect Model Performance?}
\label{subq1}

Previous efforts show that \textit{shortcut} questions widely exist in current datasets \cite{sugawara2020}.
However, there are few quantitative analysis to discuss how these \textit{shortcut} questions affect the model performance.
A reasonable guess is that, when trained with too many \textit{shortcut} questions, the models tend to fit the \textit{shortcut} tricks, which are possible solutions to a large amount of questions in training.
We thus argue that \textbf{\textit{the high proportions of \textit{shortcut} questions in training data make models rely on the \textit{shortcut} tricks.}}

One straightforward way to elaborate on this point is to observe the model performance on challenging test questions when the model is trained with different proportions of \textit{shortcut} questions. 
For example, if a model trained on a dataset, in which 90\% of questions are \textit{shortcut} ones, cannot perform as well as its 10\% variant on challenging test questions, that will probably indicate that higher proportions of \textit{shortcut} questions in the training data may hinder the model from learning other challenging skills.

\paragraph{Setup}~ We evaluate two popular MRC models, BiDAF \cite{bidaf} and BERT-base \cite{bert}, which are widely adopted in the research for \textit{shortcut} phenomena \cite{sugawara2018makes,min-etal-2019-compositional,Si2019WhatDB,sugawara2020}. 
For each combination of model and dataset, we train 10 versions of the model, adjusting the proportion of \textit{shortcut} questions in the training set from 0\% to 90\%, and report performance on pure challenging and pure \textit{shortcut} test sets. 
We report the mean and standard deviation in five runs to alleviate the impact of randomness. 
Detailed settings are elaborated in Appendix~\ref{detailsettings}.

\paragraph{Results and Analysis}~ 
Figure~\ref{bidafbert_anstype} shows the performance of BiDAF and BERT on \textit{QWM-Para} and \textit{SpM-Para} when trained with various proportions of \textit{shortcut} questions.
For both models, the F1 scores on challenging versions of both test sets drop substantially with the increase in \textit{shortcut} questions for training (Figure~\ref{bidafbert_anstype}~(a) $\sim$ (d)).
This result indicates that higher proportions of \textit{shortcut} questions in training limit the model's ability to solve challenging questions.

Take BiDAF on \textit{QWM-Para} as an example (Figure~\ref{bidafbert_anstype}~(a)).
The F1 score on the test set of challenging questions is 69\% after training BiDAF with a dataset entirely composed of challenging questions, showing that even a simple model is able to learn the paraphrasing skill from \textit{shortcut}-free training data. 
As the proportion of \textit{shortcut} training questions increases, the model tends to learn \textit{shortcut} tricks and performs worse on the challenging testing data.
The F1 score on challenging questions drops to 55\% when 90\% of the training data are  \textit{shortcut} questions.
This drop shows that training data with a high proportion of \textit{shortcuts} actually hinders the model from capturing paraphrasing skills to solve challenging questions.
In contrast, the performance on \textit{shortcut} questions are relative steady to the changes of \textit{shortcut} proportions during training.
When trained with sufficient challenging questions, 
models not only perform well on comprehension challenges, but also correctly answer the \textit{shortcut} questions where only partial evidence is required.

In Figure~\ref{bidafbert_anstype}, we can observe similar trends in model performance on \textit{SpM-Para}.
The performance on challenging questions also drops with higher proportions of \textit{shortcut} training questions.
Compared with BiDAF, although the overall scores of BERT are better, BERT also performs poorly on questions that require to perform paraphrasing when trained with more \textit{shortcut} questions, as shown in Figure~\ref{bidafbert_anstype}~(b) \& (d).

\paragraph{Case Study}~ When answering the example question \textbf{Q.4} from \textit{QWM-Para}, BiDAF trained with pure challenging questions tends to detect the correlation between \textit{graduate} and \textit{received his master's degree}, and locates the correct answer \textit{1506} when there are two spans matching the question word \textit{when}.
However, when there are more than 70\% \textit{shortcut questions} in training, BiDAF only captures the type constraint from the question word \textit{when}, and fails to identify the paraphrasing phenomenon to answer the challenging version. 

\begin{table}[h!]
\small
\centering
\begin{tabular}{p{7.2cm}}
\toprule
\textbf{Q.4}: When did Luther graduate?  \\
\textbf{P-challenging}: In 1501, at the age of 19, he entered the University of Erfurt. ... He received his master's degree in \textit{[1506]$_{Ans}$}  \\
\textbf{P-shortcut}: He received his master's degree in \textit{[1506]$_{Ans}$}  \\
\bottomrule
\end{tabular}
\end{table}

\begin{figure*}[t]
\centering
\includegraphics[width=1.\textwidth]{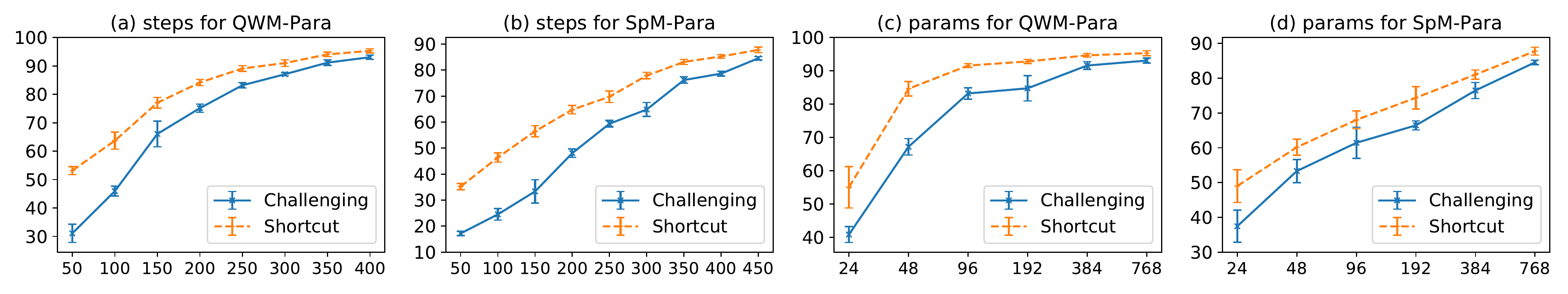}
\caption{\label{converge}
F1 scores on training sets when BERT learns challenging and \textit{shortcut} questions with different optimizing steps ((a) \& (b)) and parameter size (represented by the unmasked hidden size in the last hidden layer of BERT, (c) \& (d)).
The error bars represent the standard deviations of five runs. }
\end{figure*}

\section{Whether \textit{Question Word Matching} is Easier to Learn than \textit{Paraphrasing}?}
\label{subqnew}

It is still confusing that, with the coexistence of both \textit{shortcut} and challenging questions for training, even in a 50\%-50\% distribution, both BERT and BiDAF still learn \textit{shortcut} tricks better, thus, achieve much higher performance on \textit{shortcut questions} comparing to the challenging ones.
We think one possible reason is that \textbf{\textit{MRC models may learn the \textit{shortcut} tricks, like \textit{QWM}, with less computational resources than the comprehension challenges, like identifying paraphrasing}}. 
In this case, MRC models could better learn the \textit{shortcut} tricks with equal or even lower proportions of \textit{shortcut} questions during training.

To validate this hypothesis, we propose two simple but effective methods to measure the difference in required computational resources.
Specifically, we can train models with either pure \textit{shortcut} questions or challenging ones, and compare the learning speed and required parameter sizes when achieving certain performance levels on the training sets.

For learning speed, we train MRC models with different steps and observe how the performance changes.
Intuitively, models should converge faster on easier training data.

For parameter sizes, our intuition is that the models should learn the easier questions with fewer parameters.
However, the high computational consumption prevents us from pre-training the models like BERT with different parameter sizes.
To simulate BERT with fewer parameters, we mask some hidden units in the last hidden layer of BERT and use the number of unmasked units to reflect the parameter size. 
The information in these masked units could not be conveyed to the span boundary prediction module. Thus, only partial parameters could be used to make the final predictions.

\paragraph{Setup}~ We use BERT as the basic learner and train on the training sets of \textit{QWM-Para} and \textit{SpM-Para}.
We report model performance on the training data with various learning settings. 
We use all the parameters when adjusting learning steps.
When tuning parameter size, we fix the learning steps to 400 and 450 for \textit{QWM-Para} and \textit{SpM-Para}, respectively.
All other settings including batch size, optimizer, and learning rate are fixed.
We report the mean and standard deviation in five runs to alleviate the impact of randomness.
The implementation details are similar to \S\ref{subq1}, elaborated in Appendix~\ref{detailsettings}.

\paragraph{Results and Analysis}~ 
Figure~\ref{converge} compares the performance of BERT trained on the \textit{shortcut} questions and challenging questions separately under different settings.
On both \textit{QWM-Para} and \textit{SpM-Para}, BERT converges faster in learning \textit{shortcut} questions than learning challenging ones (Figure~\ref{converge}~(a) \& (b)).
When fixing the training steps, BERT could learn to answer the \textit{shortcut} questions with fewer parameters (Figure~\ref{converge}~(c) \& (d)). These results show that \textit{shortcut} questions may be easier for models to learn than the ones requiring complex reasoning skills.

Take \textit{QWM-Para} as an example.
As can be seen from Figure~\ref{converge}~(a), BERT trained on the \textit{shortcut} questions achieves a 90\% F1-score on the training set after 250 steps. When trained on the challenging version, this score will not reach 90\% until 350 steps. 
This result indicates that models could learn to answer the \textit{shortcut} questions with the \textit{QWM} trick faster than the paraphrasing skill. 

When we train BERT on \textit{QWM-Para} with different numbers of output units masked (Figure~\ref{converge}~(c)), BERT could 
reach the F1-score of 91\% on \textit{shortcut} data with no fewer than 96 unmasked hidden units. 
However, when trained on the challenging questions, BERT has to use 384 hidden units to reach the 91\% F1 score, which indicates that the questions with the \textit{paraphrasing} challenge may require more parameters to learn. 

We observe similar trends on
\textit{SpM-Para} (Figure~\ref{converge}~(b) \& (d)). 
BERT requires more parameters and training steps to learn the challenging version questions in \textit{SpM-Para} than the \textit{shortcut} version.
To some extent, these results confirm our hypothesis that learning to answer questions with \textit{shortcut} tricks like \textit{SpM} or \textit{QWM} requires smaller amounts of computational resources than the questions requiring challenging skills like paraphrasing.

\paragraph{Case Study}~ For the example question \textbf{Q.5}, BERT trained on \textit{shortcut} questions could correctly answer its \textit{shortcut} version and find the only location name, \textit{Palácio da Alvorada}, with only 48 unmasked hidden units. 
However, when trained with the challenging data only, the model predicts the other location name, \textit{the Monumental Axis} as the answer with such parameter size.
BERT could not recognize the paraphrasing relationship between \textit{the place the president live} and \textit{presidential residence} and choose the correct answer, \textit{Palácio da Alvorada}, from the distracting location name, until using all 748 parameters.

\begin{table}[h!]
\small
\centering
\begin{tabular}{p{7.2cm}}
\toprule
\textbf{Q.5}: Where does the president of Brazil live, in Portuguese?  \\
\textbf{P-challenging}: ... on a triangle of land jutting into the lake, is the Palace of the Dawn (\textit{[Palácio da Alvorada]$_{Ans}$}; the presidential residence). Between the federal and civic buildings on the Monumental Axis is the city 's cathedral...  \\
\textbf{P-shortcut}: ... on a triangle of land jutting into the lake, is the Palace of the Dawn (\textit{[Palácio da Alvorada]$_{Ans}$}; the presidential residence) \\
\bottomrule
\end{tabular}
\end{table}

\begin{figure*}[t!]
\centering
\includegraphics[width=1.0\textwidth]{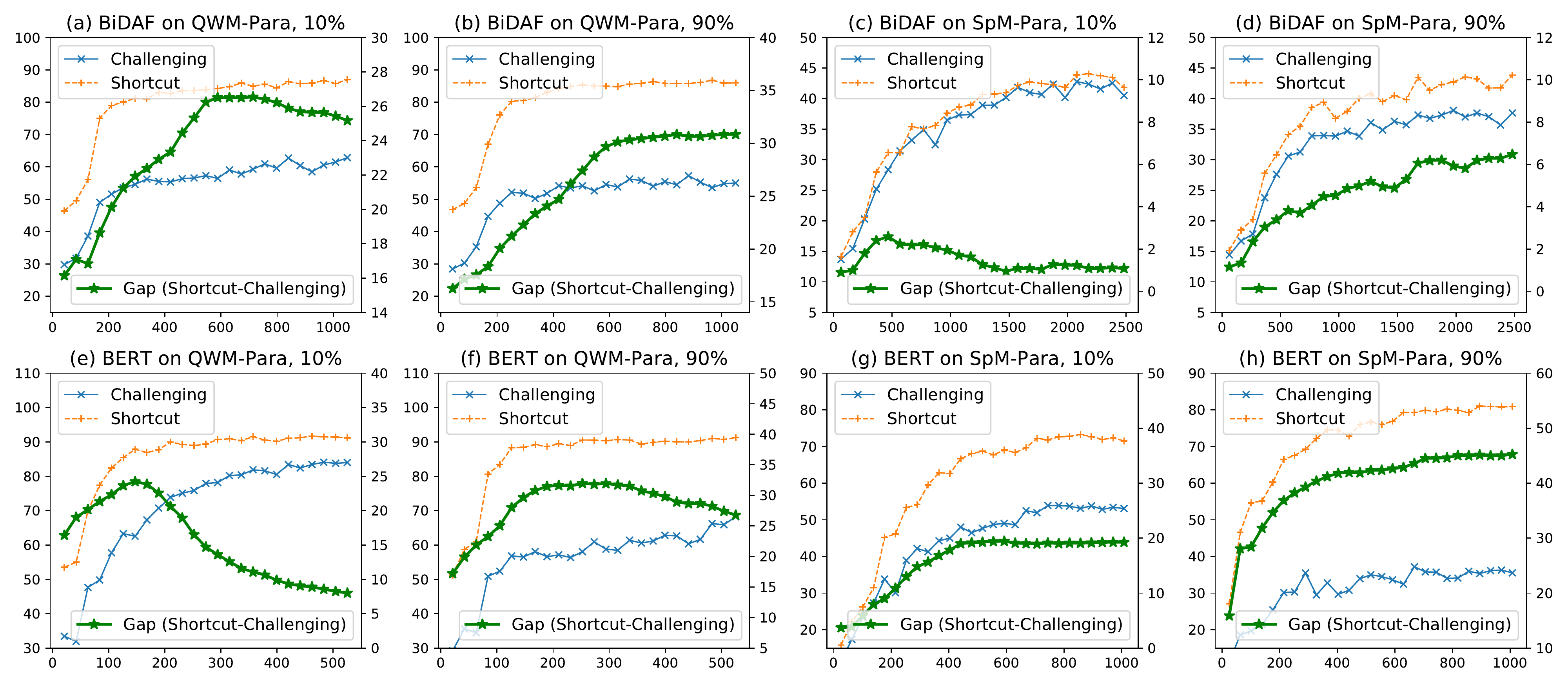}
\caption{\label{type_iter}
F1 scores on challenging and \textit{shortcut} questions with different training steps under different settings. 
10\% and 90\% are the proportions of \textit{shortcut} questions in the training datasets.
\textit{Gaps} (green lines with ``$\star$'' dots) represent the performance gap between \textit{shortcut} questions and challenging ones, which is smoothed by averaging over fixed-size windows to mitigate periodic fluctuations. 
}
\end{figure*}

\section{How do Models Learn \textit{Shortcuts}?}
\label{subq2}

In previous section, we show that \textit{shortcut} questions are \textit{easier} to learn compared to the questions that require the complex \textit{paraphrasing} skill. 
Then, it is interesting that, trained with a mixture of both versions of questions, how such discrepancy affects or even drives the learning procedure,
e.g., how the increasing of challenging training questions alleviate the over reliance on \textit{shortcut} tricks.

We guess one of the possible reasons is how most existing MRC models are optimized. 
We hypothesize that with a larger proportion of \textit{shortcut} questions for training, 
\textbf{\textit{the models have learned the shortcut tricks at the early stage, which may affect the models' further exploration for challenging skills.}}
In other words, in the early stage of training, models tend to find the easiest way to fit the training data with gradient descent.
Since the \textit{shortcut} tricks require less computational resources to learn, fitting these tricks may be a priority.
Afterwards, since the \textit{shortcut} trick can be used to answer most of the training questions correctly, the limited unsolved questions remained may not motivate the models to explore sophisticated solutions that require challenging skills.

To validate this idea, we investigate how the models converge during training with different \textit{shortcut} proportions in the training data.
Notice that if a model can only answer the \textit{shortcut} version of a question correctly, it is highly likely that the model only adopts the \textit{shortcut} trick for this question instead of performing sophisticated reasoning skills.
Thus, we think the performance gap on two versions of test data may indicate to what extent the model relies on the \textit{shortcut} tricks, e.g., the smaller the performance gap is, the stronger complex reasoning skills the model have learned.

\paragraph{Setups}~ We explore how BERT and BiDAF converge with 10\% and 90\% \textit{shortcut} training questions on \textit{QWM-Para} and \textit{SpM-Para}.
We report the F1 scores on the challenging and \textit{shortcut} test questions, respectively, and together with their performance gaps. 
We compare the model performance 
at different learning steps to investigate when and how well the models learn the \textit{shortcut} tricks and the challenging comprehension skills.
The implementation details are the same as \S\ref{subq1}, elaborated in Appendix~\ref{detailsettings}.

\paragraph{Results and Analysis}~ 
Figure~\ref{type_iter} illustrates how the MRC models converge during training under different settings.
The gap line (green with ``$\star$'') shows the gap between models' performance on  \textit{shortcut} questions and that on challenging ones.

For all settings, in the first few epochs, the model performance on \textit{shortcut} questions increases rapidly, much faster than that on challenging questions, causing a steep rise of the performance gap.
This result indicates that models may learn the \textit{shortcut} tricks at the early stage of training, thus quickly and correctly answering more \textit{shortcut} questions. 
And then, for the following epochs after reaching the peaks, the gap lines slightly go down (Figure~\ref{type_iter} (a), (c), (e), and (f)) or maintain almost unchanged (Figure~\ref{type_iter} (b), (d), (g), and (h)),
which also indicates the models may learn the challenging skills later than \textit{shortcut} tricks. 
One possible reason is the gradient based optimizer drives the model to optimize the global target greedily via the easiest direction. 
Thus, trained with a mixture of \textit{shortcut} and challenging questions, models choose to learn the \textit{shortcut} tricks, which require less computational resources to learn, earlier than the sophisticated paraphrasing skills.

Comparing models with different proportions of \textit{shortcut} training questions, we find that, with 90\% \textit{shortcut} training questions (Figure~\ref{type_iter}~(b), (d), (f) and (h)), the performance gap remains at a high level in the later training stage, where the performance on the challenging test questions is relatively lower.
These results provide evidence that, for most cases, after fitting on the \textit{shortcut} questions, models seem to fail to explore the sophisticated reasoning skills.

When there are only 10\% of \textit{shortcut} training data (Figure~\ref{type_iter}~(a), (c), (e), and (g)), we can observe that after a few hundreds of steps, the gap lines stop increasing and even slightly go down. 
This phenomenon shows that higher proportions of challenging questions in the training set could encourage the models to explore the sophisticated reasoning skills, but in a later stage of training.

Take BiDAF trained on \textit{QWM-Para} as an example (Figure~\ref{type_iter}~(a) \& (b)).
The F1 scores on \textit{shortcut} test questions increase quickly in the first 300 steps, while the performance gap also widens rapidly, indicating a possible fast fitting on the \textit{shortcut} tricks.
In Figure~\ref{type_iter}~(b), with 90\% \textit{shortcut} training questions, the model performance on challenging questions are relatively steady during the next 800 steps, while the F1 score on \textit{shortcut} questions maintains a high level of about 85\%.
This result shows that after fitting on the \textit{shortcut} tricks, the model trained with a higher \textit{shortcut} proportion has almost correctly answered all the \textit{shortcut} questions but fail to answer the challenging ones.
Actually, with the gradient based optimizer, it is difficult for the model to learn the challenging questions while keeping the high performance on the \textit{shortcut} ones, which account for 90\% of the training set.
We guess it is because the few unsolved challenging questions could not motivate the model to explore sophisticated reasoning skills.

On the contrary, when 90\% of the training data require challenging skills, the gap line peaks at 0.27, as shown in Figure~\ref{type_iter}~(a).
Afterwards, the gap decreases to 0.24, with the F1 score on challenging questions increasing to more than 60\%.
Larger proportions of challenging questions for training prevent the models from heavily relying on the \textit{shortcut} tricks.
This phenomenon may be because, with fewer \textit{shortcut} questions in training, the fitting of \textit{shortcut} tricks only benefits the training objective in a small favor.
The large number of challenging questions that can not be correctly answered during the early training steps now encourage models to explore more complicated reasoning skills.

\paragraph{Case Study}~ When answering the example question \textbf{Q.6} from \textit{SpM-Para}, BERT trained with 10\% \textit{shortcut} questions tends to learn the \textit{simple matching} trick quickly and correctly answers the \textit{shortcut} version as early as 380 steps. 
However, the model cannot correctly answer the challenging variant until 630 steps. 
This difference demonstrates that, training with both type of questions, BERT can learn the \textit{simple matching} trick earlier than identifying the required paraphrasing between \textit{why defections occur} and \textit{errors caused by}.

\begin{table}[h!]
\small
\centering
\begin{tabular}{p{7.2cm}}
\toprule
\textbf{Q.6}: Why do these defections occur?  \\
\textbf{P-challenging}: ... Most of these errors are caused by \textit{[economic or financial factors]$_{Ans}$} ... \\  
\textbf{P-shortcut}: ... Most of these defections occur because of \textit{[economic or financial factors]$_{Ans}$}. Most of these errors are caused by \textit{[economic or financial factors]$_{Ans}$}. ... \\  
\bottomrule
\end{tabular}
\end{table}

\section{Related Works}

Reading documents to answer natural language questions has drawn more and more attention in recent years \cite{xu-etal-2016-hybrid,bidaf,lai2019lattice,glass-etal-2020-span}.
Most previous works focus on revealing the  \textit{shortcut} phenomenon in MRC from different perspectives.
They find that manually designed features \cite{chen2016thorough} or simple model architectures \cite{weissenborn-etal-2017-making} could obtain competitive performance, indicating that complicated inference procedure may be dispensable.
Even without reading the entire questions or documents, models can still correctly answer a large portion of the questions \cite{sugawara2018makes,kaushik2018much,min-etal-2019-compositional}.
Therefore, current MRC datasets may lack the benchmarking capacity on requisite skills \cite{sugawara2020}, and models may be vulnerable to adversarial attacks \cite{jia2017adversarial,wallace-etal-2019-universal,Si2019WhatDB}.
However, they do not formally discuss or analyze why models could learn \textit{shortcuts} 
from the perspectives of the learning procedure.

On the way of designing better MRC datasets, \citet{jiang-bansal-2019-avoiding} construct adversarial questions to guide model learning the multi-hop reasoning skills. \citet{Bartolo2020BeatTA} propose a model-in-loop paradigm to annotate challenging questions.
More recent works \cite{jhamtani-clark-2020-learning,ho-etal-2020-constructing} propose new datasets with evidence based metrics to evaluate whether the questions are solved via \textit{shortcuts}.
Our work aims at providing empirical evidence and analysis to the community by tracing into the learning procedure and explaining how the MRC models learn \textit{shortcuts}, which is orthogonal to the existing works.

For a more general machine learning perspective, there are also efforts trying to explain how models learn easy and hard instances during training. \citet{kalimeris2019sgd} prove that models tend to learn easier decision boundaries at the beginning stage of training.
Our results empirically confirms this theoretical conclusion in the task of MRC and quantitatively explain that larger proportions of \textit{shortcut} questions in training make MRC models rely on \textit{shortcut} tricks, rather than comprehension skills like recognizing the paraphrase relationship.

\section{Conclusions}

In this work, we try to answer why many MRC models learn \textit{shortcut} tricks while ignoring the pre-designed comprehension challenges that are purposely embedded in many benchmark datasets.
We argue that \textbf{\textit{large proportions of \textit{shortcut} questions in training data push MRC models to rely on \textit{shortcut} tricks excessively}}. 
To properly investigate, we first design two synthetic datasets where each instance has a \textit{shortcut} version paired with a challenging one which requires \textit{paraphrasing}, a complex reasoning skill, to answer, rather than performing \textit{question word matching} or \textit{simple matching}. 
With these datasets, we are able to adjust the proportion of \textit{shortcut} questions in both training and testing, while maintaining other factors relatively steady.
We propose two methods to examine the model training process regarding the \textit{shortcut} questions, which enable us to take a closer look at the learning mechanisms of BiDAF and BERT under different training settings.
We find that learning \textit{shortcut} questions generally requires less computational resources, 
and MRC models usually learn the \textit{shortcut} questions at their early stage of training.
Our findings reveal that, with larger proportions of \textit{shortcut} questions for training, 
MRC models will learn the \textit{shortcut} tricks quickly while ignoring the designed comprehension challenges, since the remaining truly challenging questions, usually limited in size, may not motivate models to explore sophisticated solutions in the later training stage.

\section*{Acknowledgments}
This work is supported in part by the National Hi-Tech R\&D Program of China (No.2020AAA0106600), the NSFC under grant agreements (61672057, 61672058). For any correspondence, please contact Yansong Feng.


\bibliographystyle{acl_natbib}
\bibliography{acl2021}

\newpage

\begin{appendix}
\section{Implement Details}
\label{detailsettings}

\paragraph{Synthetic Dataset Construction}~ During the construction of synthetic datasets, we used Stanford CoreNLP \cite{stanfordcorenlp} to identify named entities and stop-words. 

We set two empirical thresholds for identifying questions can be solved by \textit{Simple Matching} or requiring paraphrasing skills.
We consider a question as solvable via the \textit{simple matching} trick if more than 75\% of non-stop words in the question exactly appear in the answer sentence. 
On the other hand, if the matching rate is below 25\%, we think it is unsolvable via \textit{simple matching}, calling for other skills like paraphrasing. 
Thus, in dataset construction, if the matching rate is above 25\% after paraphrasing, we consider that the back translation fails to incorporate paraphrasing skills into the instance.

We construct the synthetic datasets from SQuAD \cite{squad}.
Compared with more recent MRC datasets \cite{hotpot,kwiatkowski-etal-2019-natural}, most questions in SQuAD can be solved by a single sentence with \textit{simple matching} so that we can conveniently use back translation to construct questions with paraphrasing challenges.

\paragraph{Paraphrasing in \textit{SpM-para}}

When constructing the \textit{SpM-Para} dataset, we only select the instances whose questions are very similar to the corresponding answer sentences (overlap $>$ 75\%) to ensure that a \textit{simple matching} step can obtain the answers. For the \textit{shortcut}-version of an instance, we  
insert the paraphrased answer sentence into passage and
keep both the original answer sentence and paraphrased answer sentence (see Algorithm~\ref{algo_spm}).
This operation aims to control the \textit{shortcut} instances to have both \textit{shortcut} solutions and challenging solutions.
For the challenging version, we only keep the paraphrased answer sentences in the passages and discard the original answer sentence, so that such instances can only be solved by identifying the embedded paraphrasing relationship.

\paragraph{Hyper-Parameters for QA models}~ 
We adopt BERT (BERT-based uncased) \cite{bert} and BiDAF \cite{bidaf} models with the implementation in SogouMRC tools \cite{wu2019sogou}.
The hyper-parameters are shown in Table~\ref{hyper-parameters}. 
We used 100-d glove vectors \cite{glove} in BiDAF.
Notice that these hyper-parameters are adopted in \S3, \S4, and \S5.
Our code and datasets can be found in \url{https://github.com/luciusssss/why-learn-shortcut}

For the \textit{simple matching} setting where multiple answer spans may appear in one passage, we follow \cite{pang2019has} and aggregate the possibilities of each span before computing the likelihood losses. 

\paragraph{Data Sampling in Difficulty Evaluation}~ In \S4, we train BERT on the training sets of \textit{QWM-Para} and \textit{SpM-Para} and observe how the model converges with different learning steps and parameter sizes.
However, we find BERT achieves outstanding performance on both datasets with only one or two epochs.
This is because the strong learning ability of BERT model and, with only one kind of answering pattern, both the pure \textit{shortcut} and challenging training sets are relatively easy to learn.

Under this circumstance, BERT performance on most of the evaluation checkpoints after one epoch will almost reach the final performance, which make the comparison vague. If we compare the checkpoints within one epoch, considering that models have only been trained on partial training data, the evaluation results would reflect the models' generalization ability on unseen questions. 
This differs from our purpose of evaluation, namely comparing the fitting difficulty of different kinds of questions.
Therefore, we randomly sample 1000 pair of instances for training and evaluation.
With less training data, BERT will not converge in only one or two epochs, thus we could truly evaluate the learning ability.

\paragraph{Computation Cost}~ We train the models on an NVIDIA 1080 Ti GPU. The number of parameters is 5M for BiDAF and 110M for BERT. The average training time on synthetic datasets for an epoch is 1 minute for BiDAF and 10 minutes for BERT. 

\begin{table}[ht]
\small
\centering
\begin{tabular}{lll}
\toprule
              &BERT-base Uncased                                                                                      & BiDAF                                                                                                                                                \\ \midrule
\# Epoch         & 3                                                                                         & 15                                                                                                                                                   \\
Batch size     & 6                                                                                         & 30                                                                                                                                                   \\
Optimizer     &
AdamWeightDecay & Adam
\\
Learning rate & 3e-5 &  1e-3    \\
\bottomrule
\end{tabular}
\caption{
\label{hyper-parameters}
Hyper-Parameters for the experiments in \S4, \S5, and \S6.
}
\end{table}

\section{A Variant of \textit{QWM-Para} Dataset}

When we train models with different proportions of \textit{shortcut} questions on \textit{QWM-Para} (Figure~3~(a) \& (b), which is described in \S3), we observe that even with pure challenging questions in training, BiDAF and BERT still perform much better on \textit{shortcut} questions than on the challenging ones.
We think this is possibly because in these settings, models fail to exploit the paraphrasing skill but learn to guess one from the the entities matching the question words as the answer.
Using such a guessing trick instead of the paraphrasing skill could improve the performance on the challenging questions to some extent, but it results in more  gains on the \textit{shortcut} questions. 
Therefore, even with 100\% challenging questions in training, the gap between the performance on challenging and \textit{shortcut} test questions is still wide.

To avoid these guess solutions, we redesign a variant of the \textit{QWM-Para} dataset, named as \textit{\textbf{QWM/subs-Para}}.\footnote{\textit{subs} refers to substituted, elaborated in \S B.1.}
We aim at investigating:
1) Whether this variant could avoid the guessing alternative and decrease the performance gaps between challenge questions and \textit{shortcut} ones when training with relatively lower \textit{shortcut} proportions.
2) Whether the experiments on this variant still confirms our previous findings about how \textit{shortcut} questions in training affect model performance and learning procedure, as described in \S3 and \S5. 

\begin{figure}[t!]
    \centering
    \includegraphics[width=0.4\textwidth]{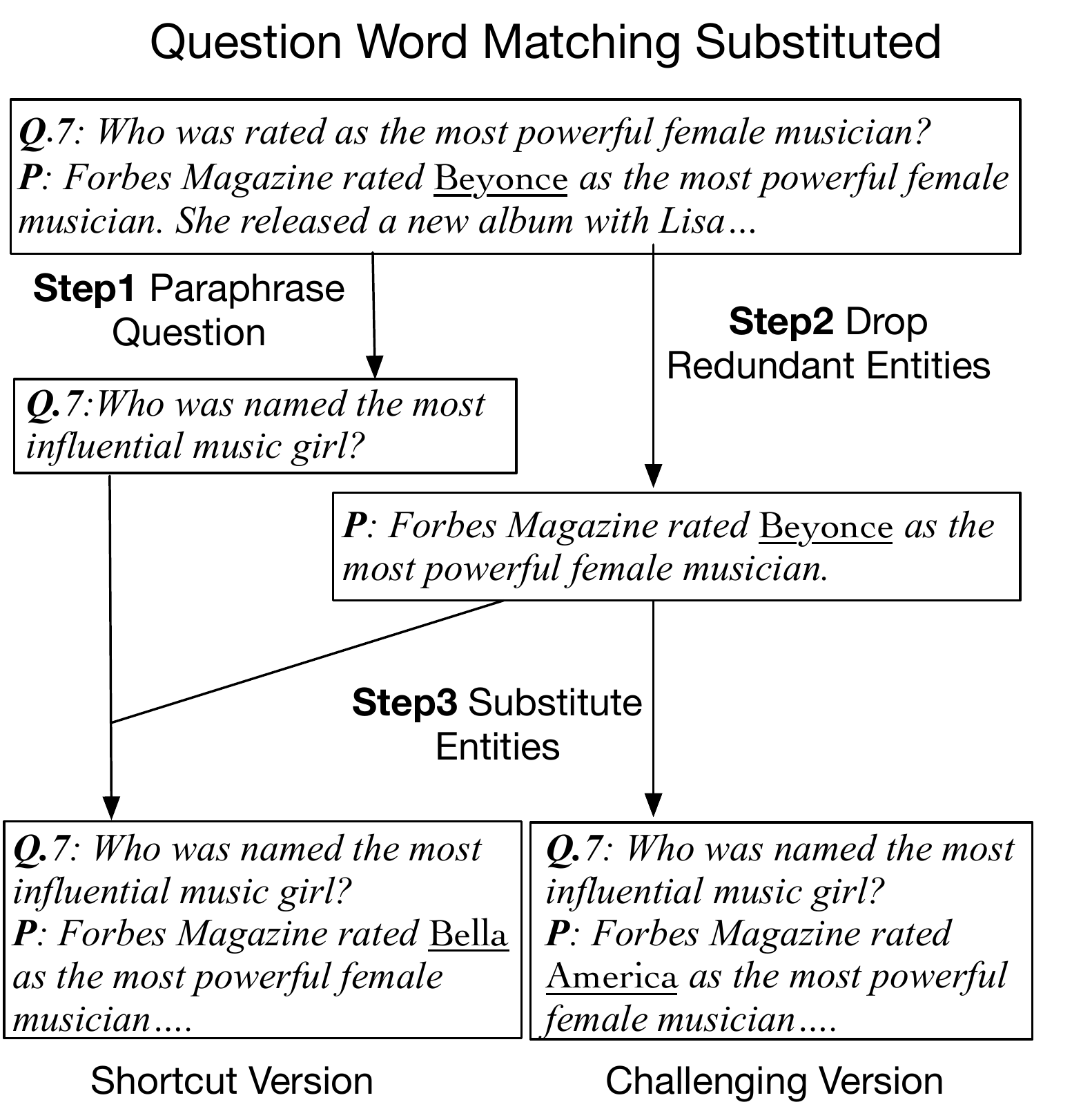}
    \caption{An illustration of how the questions in the new synthetic datasets, \textit{question word matching substituted}, are constructed from original questions.
    \label{fig:appcon}
    }
\end{figure}

\subsection{Dataset Construction}

The construction process of \textit{QWM/subs-Para} is shown in Figure \ref{fig:appcon}. 
The first two steps, question paraphrasing and redundant entities dropping, are the same as those in the construction of \textit{shortcut} questions in \textit{QWM-Para} (see \S2).
Then, we perform entity substitution to avoid the potential guessing solutions.

\begin{figure}[t!]
\centering
\includegraphics[width=0.5\textwidth]{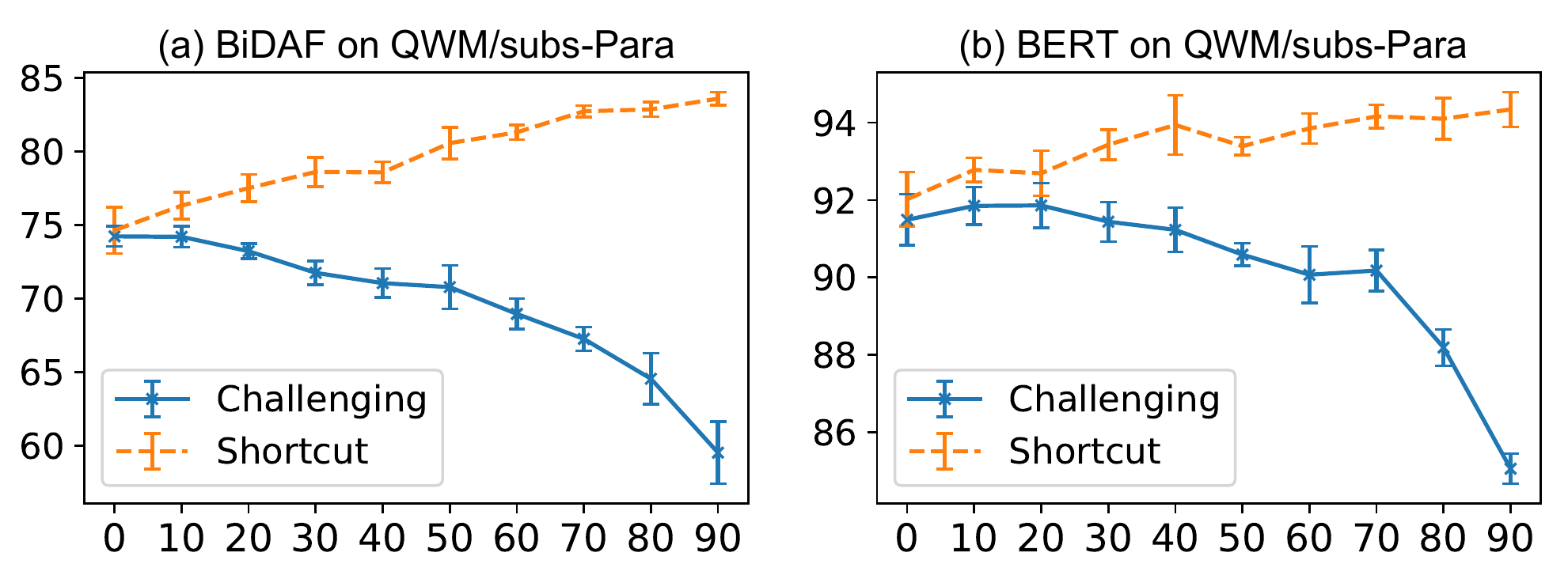}
\caption{\label{app:bidafbert_anstype}
F1 scores on challenging and \textit{shortcut} questions with different proportions of \textit{shortcut} questions in training.
This experiment is conducted on \textit{QWM/subs-Para}.
The error bars represent the standard deviations of five runs.
}
\end{figure}

\begin{figure}[t!]
\centering
\includegraphics[width=0.48\textwidth]{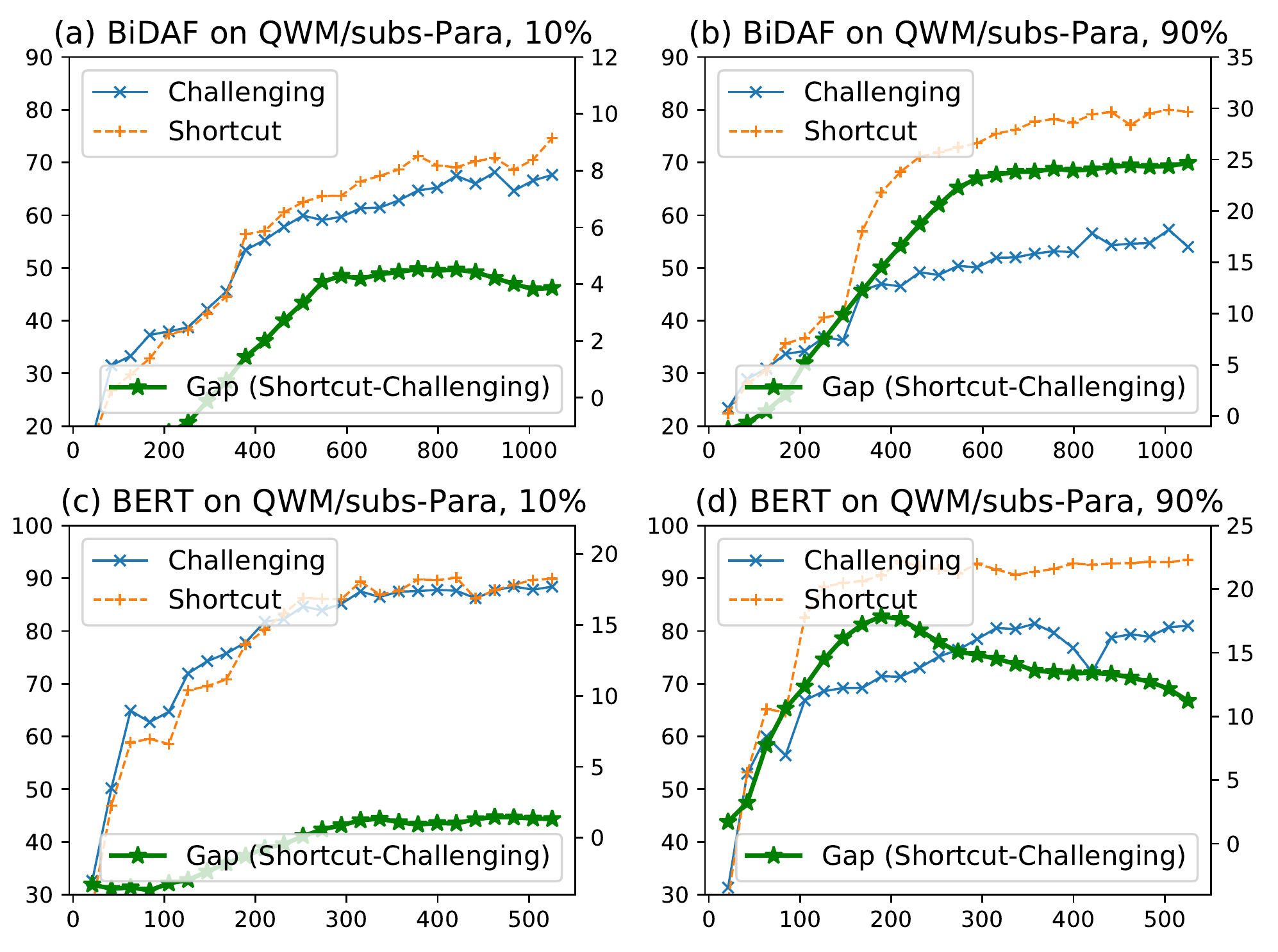}
\caption{\label{app:type_iter}
F1 scores on challenging and \textit{shortcut} questions with different training steps under different settings. 
This experiment is conducted on  \textit{QWM/subs-Para}.
10\% and 90\% are the ratios of \textit{shortcut} questions in the training datasets.
\textit{Gaps} (green lines with ``$\star$'' dots) represents the performance gap between \textit{shortcut} questions and challenging ones, which is smoothed by averaging over fixed-size windows to mitigate periodic fluctuations. 
}
\end{figure}

Particularly, for each candidate question, we substitute all the entities in the passage with random ones whose type uniformly distributes in Person/Time/Location to construct the challenging questions.
With this random substitution, one can hardly guess the correct answer via the question words.
As shown in \textbf{Q.7}, after substituting the answer entity \textit{Beyonce} to \textit{America}, one can not answer the new question by simply finding a \textit{Person} entity according to the question word \textit{who}. 
Replacing a person’s name with a location may break the original semantic, but it will force the model to comprehend the context to find the answers.
For the \textit{shortcut} version, we also conduct the random entity substitution, but within the same entity types, e.g., from \textit{Beyonce} to \textit{Bella}.

This strategy could avoid the models from learning the trick that identifying replaced words as the answers.

\subsection{Results and Analysis}

Shown in Figure~\ref{app:bidafbert_anstype}, we can see that, when constructing the challenging questions with entity substitution, both BiDAF and BERT model perform comparably between challenging and \textit{shortcut} test questions with 100\% challenging questions in training.
These results provide evidence that, after the substitution, models could not use guessing as an alternative solutions to the paraphrasing skill. 

We conduct similar experiments in \S3, \S5 on \textit{QWM/subs-Para}, which is shown in Figure~\ref{app:bidafbert_anstype} and Figure~\ref{app:type_iter}, respectively.
The tendency could also support our previous findings. 
For example, the larger \textit{shortcut} ratio expands the performance gaps between challenging and \textit{shortcut} questions in Figure~\ref{app:bidafbert_anstype}.

\end{appendix}

\end{document}